\documentclass[10pt,twocolumn,letterpaper]{article}

\usepackage[pagenumbers]{cvpr}

\usepackage{times}
\usepackage{epsfig}
\usepackage{amsmath}
\usepackage{amssymb}
\usepackage{microtype}
\usepackage{enumitem}
\usepackage[accsupp]{axessibility}
\usepackage{makecell}
\usepackage{caption}
\usepackage{pifont}

\usepackage{url}
\usepackage[table]{xcolor}
\usepackage{graphicx,wrapfig,lipsum}
\usepackage{threeparttable}
\usepackage{tabularx}
\usepackage{booktabs}
\usepackage{multirow}
\usepackage{array}
\usepackage{colortbl}
\newcolumntype{P}[1]{>{\raggedright\arraybackslash}p{#1}}

\definecolor{cvprblue}{rgb}{0.21,0.49,0.74}
\usepackage[breaklinks,colorlinks,citecolor=cvprblue]{hyperref}

\setitemize{noitemsep,topsep=0pt,parsep=0pt,partopsep=0pt}

\newcommand{\cmark}{\ding{51}}

\def\paperID{}
\def\confName{CVPR}
\def\confYear{2026}

\title{STS-Mixer: Spatio-Temporal-Spectral Mixer for 4D Point Cloud \\ Video Understanding}

\author{
Wenhao Li\textsuperscript{1} \quad
Xueying Jiang\textsuperscript{1} \quad
Gongjie Zhang\textsuperscript{2} \quad
Xiaoqin Zhang\textsuperscript{3}\quad
Ling Shao\textsuperscript{4} \quad
Shijian Lu\textsuperscript{1}\thanks{Corresponding Author.}
\\
\textsuperscript{1}{\normalsize Nanyang Technological University} \quad
\textsuperscript{2}{\normalsize Alibaba Group}  \\
\textsuperscript{3}{\normalsize Zhejiang University of Technology} \quad
\textsuperscript{4}{\normalsize University of Chinese Academy of Sciences} \\
}

\begin{document}
\maketitle

\begin{abstract}
4D point cloud videos capture rich spatial and temporal dynamics of scenes which possess unique values in various 4D understanding tasks. However, most existing methods work in the spatiotemporal domain where the underlying geometric characteristics of 4D point cloud videos are hard to capture, leading to degraded representation learning and understanding of 4D point cloud videos. We address the above challenge from a complementary spectral perspective. By transforming 4D point cloud videos into graph spectral signals, we can decompose them into multiple frequency bands each of which captures distinct geometric structures of point cloud videos. Our spectral analysis reveals that the decomposed low-frequency signals capture more coarse shapes while high-frequency signals encode more fine-grained geometry details. Building on these observations, we design Spatio-Temporal-Spectral Mixer (STS-Mixer), a unified framework that mixes spatial, temporal, and spectral representations of point cloud videos. STS-Mixer integrates multi-band delineated spectral signals with spatiotemporal information to capture rich geometries and temporal dynamics, while enabling fine-grained and holistic understanding of 4D point cloud videos. Extensive experiments show that STS-Mixer achieves superior performance consistently across multiple widely adopted benchmarks on both 3D action recognition and 4D semantic segmentation tasks. Code and models are available at \url{https://github.com/Vegetebird/STS-Mixer}. 
\end{abstract}

\section{Introduction}

4D point cloud videos comprise sequences of 3D coordinate sets that capture 3D spatial and 1D temporal information concurrently. 
They faithfully represent geometric structures and temporal dynamics of the 4D physical world and have been widely explored in various 4D perception tasks \cite{wangpvnext,jing2024x4d,shen2023pointcmp,sheng2023point}. 
Unlike structured RGB videos, point cloud videos are inherently irregular and unordered in the spatial space while maintaining regularity and order in the temporal space. 
Such asymmetry along spatial and temporal dimensions presents unique challenges for effective modeling of spatiotemporal dynamics of point cloud videos. 

\begin{figure}[t]
\centering
\includegraphics[width=1.00\linewidth]{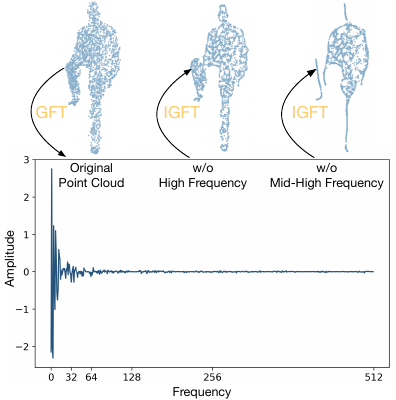}
\caption
{
Graph spectral analysis of point clouds with Graph Fourier Transform (GFT) and inverse GFT (IGFT). 
The high-frequency bands of point clouds encode fine-grained geometric details which can be observed by the difference between the original point cloud and the one w/o high frequency. 
The low-frequency bands capture most energy as well as the coarse shapes which can be observed by the similarity between the original point cloud and the one w/o mid-high frequency. 
}
\label{fig:moti}
\end{figure}

Recent studies \cite{fan2021point,fan2022point,wen2022point} explore 4D convolutional neural networks (CNNs) and Transformers to model short-term and long-term spatiotemporal dynamics of point cloud videos. 
They adopt 4D point convolutions to encode local spatiotemporal structures, and Transformers to capture global context across video frames. 
However, these studies often struggle to capture the underlying geometric characteristics of 4D point cloud videos that comprise rich abstract and global-local contexts \cite{hu2022exploring}, leading to mediocre performance in various 4D understanding tasks with point cloud videos. 

We explore spectral representations of point clouds that capture supplementary and useful geometric features for various 4D point cloud understanding tasks. 
The idea is that the point cloud data can be decomposed into multiple distinct frequency bands in the spectral space, each explicitly capturing specific geometric structures of point clouds such as global shapes and fine-grained local details \cite{zhang2014point,hu2021graph}. 
On the other hand, learning such spectral representations is a nontrivial task due to the complex spatiotemporal structure of point cloud videos. 
Specifically, unlike regularly ordered pixels in RGB images that can be transformed via Discrete Cosine Transform (DCT), points in point cloud videos are spatially unordered and irregular and cannot be well transformed with spatial-space operators like DCT. 
We thus exploit Graph Fourier Transform (GFT) to derive spectral representations of point cloud signals since graphs are naturally suitable to represent the irregular and unordered points. 

The GFT-transformed spectral representations capture well-decomposed information.
As Figure~\ref{fig:moti} shows, the low-frequency bands have much higher amplitudes than the high-frequency ones in the spectral point-cloud representation, indicating that the low-frequency bands capture the major energy and contents of point clouds. 
We perform band rejection to examine how different frequency bands capture different geometric structures. It can be observed that the reconstructed point clouds via inverse GFT (IGFT) lack fine-grained details when high-frequency bands are removed. 
Further removing the mid-frequency bands leads to more missing details though the global shapes of objects remain clear. 
The band rejection analysis reveals that each frequency band captures specific geometric structures of point clouds: high-frequency bands capture more fine-grained geometric details while low-frequency bands capture more global shapes. 
Such findings are well aligned with prior studies \cite{rosman2013patch,zhang2020hypergraph,chen2017fast} showing that low-frequency bands are more suitable for point cloud denoising while high-frequency bands are more suitable for handling redundant information. 

Leveraging the above graph spectral analysis, we design Spatio-Temporal-Spectral Mixer (STS-Mixer), a unified framework that mixes spatiotemporal modeling with frequency-aware spectral representations to capture more comprehensive geometries and temporal dynamics of point cloud videos. 
STS-Mixer represents point clouds as graphs by connecting each point to its local neighbors and transforms point coordinates into the spectral space via GFT. 
The produced GFT coefficients can be partitioned into multiple frequency bands, enabling frequency-wise decomposition of geometric structures of point clouds. 
Each decomposed frequency band can be transformed back to the spatial space via IGFT, leading to band-specific reconstructions that capture geometric structures of different scales. 
Then, our STS-Mixer blocks process such delineated multi-band signals to model both global and fine-grained geometric structures of point clouds. 
It consists of two key designs, namely, Frequency-Aware Attention (FA-Attention) and Frequency-Mixing MLP (FM-MLP). 
FA-Attention refines the spectral representation of each decomposed frequency band independently while FM-MLP facilitates information exchange across the spectral representations of different frequency bands, enabling intra-band refinement and inter-band interaction and mutual enhancement. 
Extensive experiments over both action recognition and semantic segmentation tasks show that STS-Mixer achieves superior performance consistently across multiple widely adopted benchmarks. 

The contributions of this work can be summarized in three major aspects. 
First, we examine point cloud videos via GFT and identify that each frequency band captures distinct geometric characteristics from global shapes to local details. 
To the best of our knowledge, this is the first work that explores spectral representations for 4D point cloud video understanding. 
Second, we design the STS-Mixer that decomposes point cloud videos into multiple frequency bands and dynamically integrates multi-band spectral signals with spatiotemporal information, while enabling fine-grained and holistic representation learning of point cloud videos. 
Third, extensive experiments over multiple benchmarks show that STS-Mixer achieves superior performance consistently across 3D action recognition and 4D semantic segmentation tasks. 

\begin{figure*}[t]
\centering
\includegraphics[width=1.00\linewidth]{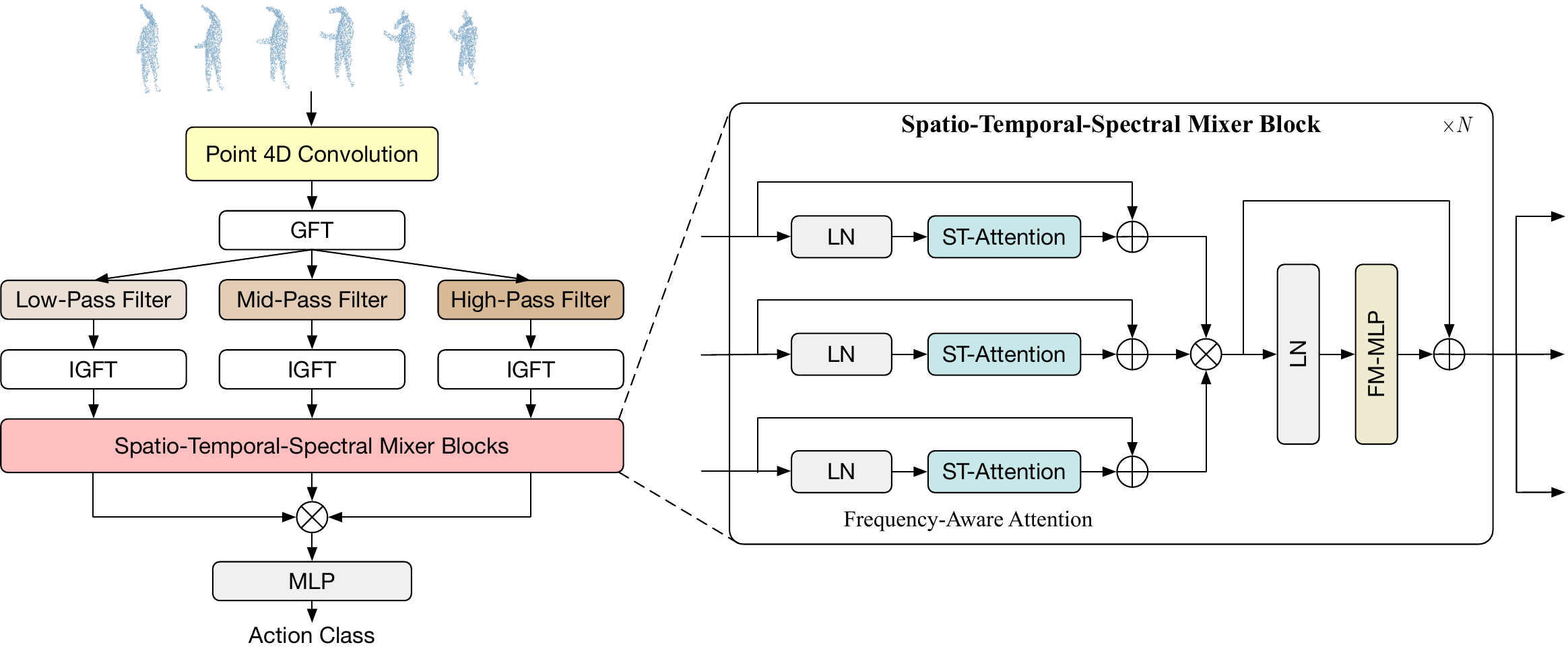}
\caption
{
The framework of the proposed STS-Mixer. 
STS-Mixer decomposes 4D point cloud videos into multi-band signals and processes them to better capture information across spatial, spectral, and temporal spaces. 
Each STS-Mixer block comes with Frequency-Aware Attention (FA-Attention) and Frequency-Mixing MLP (FM-MLP), the former refines the spectral representation of each individual frequency band independently and the latter facilitates information exchange across the spectral representations of different frequency bands for mutual enhancement. 
}
\label{fig:overview}
\end{figure*}

\section{Related Work}
\paragraph{Static Point Cloud Learning.}
Learning point clouds of a single frame has achieved remarkable progress thanks to the recent advances in deep learning techniques \cite{choe2022pointmixer,ma2022rethinking,liang2024pointmamba,zhang2025point}. 
For example, PointNet \cite{qi2017pointnet} is the pioneering work that provides an end-to-end solution that directly employs point clouds in various tasks, including 3D shape classification, part segmentation, and scene semantic parsing.
The follow-ups PointNet++ \cite{qi2017pointnetplus} and several other studies \cite{duan2019structural,yang2019modeling} further capture the local structural features of point clouds. 
Despite their promising performance, information from a single point cloud frame is inherently limited and cannot adequately capture the dynamic 4D world, particularly in scenarios involving motion. 

\paragraph{Dynamic Point Cloud Understanding.}
Several recent studies \cite{liu2023leaf,sheng2023contrastive,zhang2023complete} focus on the understanding of point cloud videos, which provide rich temporal cues and are more informative for dynamic scene analysis. 
For example, MeteorNet \cite{liu2019meteornet} extends PointNet++ \cite{qi2017pointnetplus} to videos to process raw point cloud sequences and performs explicit tracking of point motions. 
PointRNN \cite{fan2019pointrnn} introduces point-based recurrent neural networks tailored for point cloud video modeling. 
PSTNet \cite{fan2022pstnet} disentangles space and time in point cloud sequences and proposes a point spatiotemporal convolution. 
P4Transformer \cite{fan2021point} is the first to introduce a Transformer-based architecture for modeling spatiotemporal relationships in point cloud videos. 
PPTr \cite{wen2022point} uses primitive planes within the Transformer framework to capture long-term spatiotemporal dependencies. 
PST-Transformer \cite{fan2022point} further advances this line of work by employing video-level self-attention to adaptively search for relevant points across frames. 
However, these methods primarily focus on modeling the point cloud video in the spatiotemporal domain, while ignoring the importance of the spectral domain. 

\paragraph{Spectral Representation Learning for Point Clouds.}
In the spectral domain, the coarse shapes of point clouds are encoded into low-frequency components, which are suitable for denoising point clouds. 
In contrast, high-frequency components typically represent fine-grained details of point clouds, which are useful for processing redundant information. 
Several recent \cite{hu2022exploring,ramasinghe2020spectral,liang2024parameter,wen2024pointwavelet} works utilize spectral representations for static point clouds. 
For instance, GSDA \cite{hu2022exploring} perturbs point clouds in the graph spectral domain to generate adversarial examples. 
Spectral-GAN \cite{ramasinghe2020spectral} proposes a spectral-domain Generative Adversarial Network to synthesize 3D point clouds. 
PointGST \cite{liang2024parameter} fine-tunes the parameters of pre-trained point cloud models in the spectral domain. 
PointWavelet \cite{wen2024pointwavelet} employs the graph wavelet transform to capture multi-scale relationships among spectral features. 
However, these methods only consider static point cloud modeling and dynamic point cloud understanding with spectral representation has not been explored. 
In this work, we take the first step toward adopting a spectral-domain representation for 4D point cloud video understanding, revealing how frequency decomposition can complement spatiotemporal modeling. 
We decompose the point cloud sequence into multiple frequency representations and dynamically integrate them with spatiotemporal information to capture rich spatial geometries and temporal dynamics of point cloud videos. 

\section{Method}

The overview of the proposed Spatio-Temporal-Spectral Mixer (STS-Mixer) is illustrated in Figure~\ref{fig:overview}. 
Given a point cloud video $\{\left[P; F\right], P \in \mathbb{R}^{T \times N \times 3}, F \in \mathbb{R}^{T \times N \times C}\}$, where $P$ and $F$ represent the point coordinates and features respectively, and $T$ and $N$ represent the sequence length and number of points, we apply a point 4D convolution \cite{fan2021point} to encode spatiotemporal local and subsample points, resulting in features $\{\left[P^{\prime}; F^{\prime}\right], P^{\prime} \in \mathbb{R}^{T^{\prime} \times N^{\prime} \times 3}, F^{\prime} \in \mathbb{R}^{T^{\prime} \times N^{\prime} \times C}\}$. 
The sampled points $P^{\prime}$ are then transformed into the spectral domain via Graph Fourier Transform (GFT) and multiple frequency components are extracted by applying a set of spectral filters. 
Subsequently, each frequency component is individually transformed back to the data domain using inverse GFT (IGFT). 
This results in three distinct point clouds corresponding to low-frequency $P^{\prime}_{l}$, mid-frequency $P^{\prime}_{m}$, and high-frequency $P^{\prime}_{h}$ components. 
The point features $F^{\prime}$, combined with the multi-band spectral point clouds $(X_{l} = \left[P^{\prime}_{l}; F^{\prime}\right], X_{m} =\left[P^{\prime}_{m}; F^{\prime}\right], X_{h} = \left[P^{\prime}_{h}; F^{\prime}\right])$, are then fed into the STS-Mixer blocks to capture information across spatial, spectral, and temporal spaces.  
Finally, a multi-layer perceptron (MLP) maps the learned features to point-wise predictions such as action classes or semantic labels. 

\subsection{Preliminaries}

In this subsection, we give a brief description of the preliminaries in Transformer and Graph Fourier Transform (GFT). 

\noindent \textbf{Transformer.}
Transformers have achieved remarkable success in various computer vision tasks \cite{vit,mhformer,swin,hot}, owing to their strong capability in modeling long-range dependencies. 
A standard Transformer block consists of a Multi-Head Self-Attention (MSA) mechanism followed by a Multi-Layer Perceptron (MLP).

In MSA, the inputs $x \in \mathbb{R}^{n \times d}$ are linearly projected into queries $Q \in \mathbb{R}^{n \times d}$, keys $K \in \mathbb{R}^{n \times d}$, and values $V \in \mathbb{R}^{n \times d}$, where $n$ is the sequence length and $d$ is the feature dimension. 
The scaled dot-product attention is computed as: 
\begin{equation}
  \operatorname{Attention}(Q, K, V)=\operatorname{Softmax}\left(\frac{{Q K^{T}}}{\sqrt{d}}\right) V.
\end{equation}

The queries, keys, and values are split into $h$ heads, and attention is performed in parallel across these heads.
The outputs from all heads are then concatenated. 

The MLP consists of two linear layers for non-linear transformation:
\begin{equation}
  \label{equ:mlp}
  \operatorname{MLP}(x)=\sigma\left(x W_{1}+b_{1}\right) W_{2}+b_{2},
\end{equation}
where $\sigma(\cdot)$ denotes the GELU activation function, $W_{1} \in \mathbb{R}^{d \times d_{m}}$ and $W_{2} \in \mathbb{R}^{d_{m} \times d}$ are the weights of the two linear layers, and $b_{1} \in \mathbb{R}^{d_{m}}$ and $b_{2} \in \mathbb{R}^{d}$ are the corresponding biases. 

Let $X_{\ell-1} \in \mathbb{R}^{N \times D}$ be an input feature, and a Transformer block can be calculated as:
\begin{align}
X^{\prime}_{\ell} &= \operatorname{Attention}(\operatorname{LN}(X_{\ell-1})) + X_{\ell-1}, 
\label{equ:mlpmixer_spatial} \\
X_{\ell} &= \operatorname{MLP}(\operatorname{LN}(X^{\prime}_{\ell})) + X^{\prime}_{\ell},
\label{equ:mlpmixer_channel}
\end{align}
where $\operatorname{LN}(\cdot)$ denotes the LayerNorm \cite{ba2016layer}. 

\noindent \textbf{Graph Fourier Transform.}
Let $\mathcal{G}=\{\mathcal{V}, \mathcal{E}, \boldsymbol{W}\}$ denote a graph, where $\mathcal{V}$ represents the set of vertices, $\mathcal{E}$ is the edges, and $\boldsymbol{W}$ is the adjacency matrix. 
The graph Laplacian matrix \cite{shuman2013emerging} is defined as $\boldsymbol{L} = \boldsymbol{D} - \boldsymbol{W}$, where $\boldsymbol{D}$ is the diagonal matrix with each element $\boldsymbol{D}_{i, i}=\sum_{j=0}^{n-1} \boldsymbol{W}_{i, j}$. 
The eigen-decomposition can be calculated by $\boldsymbol{L} = \boldsymbol{U} \boldsymbol{\Lambda} \boldsymbol{U}^\top$, where $\boldsymbol{U}=\left[\boldsymbol{u}_1, \ldots, \boldsymbol{u}_n\right]$ is an orthonormal matrix composed of eigenvectors $\mathbf{u}_i$, and eigenvalues $\boldsymbol{\Lambda} = \mathrm{diag}(\lambda_0, \ldots, \lambda_{n-1})$ are referred as the graph frequencies. 

Given a graph signal $X \in \mathbb{R}^{n \times c}$, its GFT coefficients can be defined as \cite{hammond2011wavelets}:
\begin{equation}
    \hat{X} = \operatorname{GFT}(X) = \boldsymbol{U}^\top X. 
\end{equation}

Similarly, the inverse Graph Fourier Transform (IGFT) is then defined as:
\begin{equation}
X = \operatorname{IGFT}(\hat{X}) = \boldsymbol{U} \hat{X}.
\end{equation}

\subsection{Transform into Spectral Domain}

We observe that most existing point cloud video methods \cite{fan2021point,fan2022point,wen2022point} directly process point data in the spatiotemporal domain. 
However, these approaches cannot well capture the underlying geometric characteristics of point clouds. 
To address such an issue, we explore point cloud video understanding in the spectral domain. 
Unlike images that lie on regular grids, point clouds are defined over irregular domains without inherent ordering, making traditional transformations such as the DCT unsuitable \cite{hu2022exploring,liang2024parameter}. 
Hence, we employ GFT to transform the point cloud into the spectral domain, which is more appropriate for processing irregular data. 

\begin{figure}[t]
\includegraphics[width=1.0\linewidth]{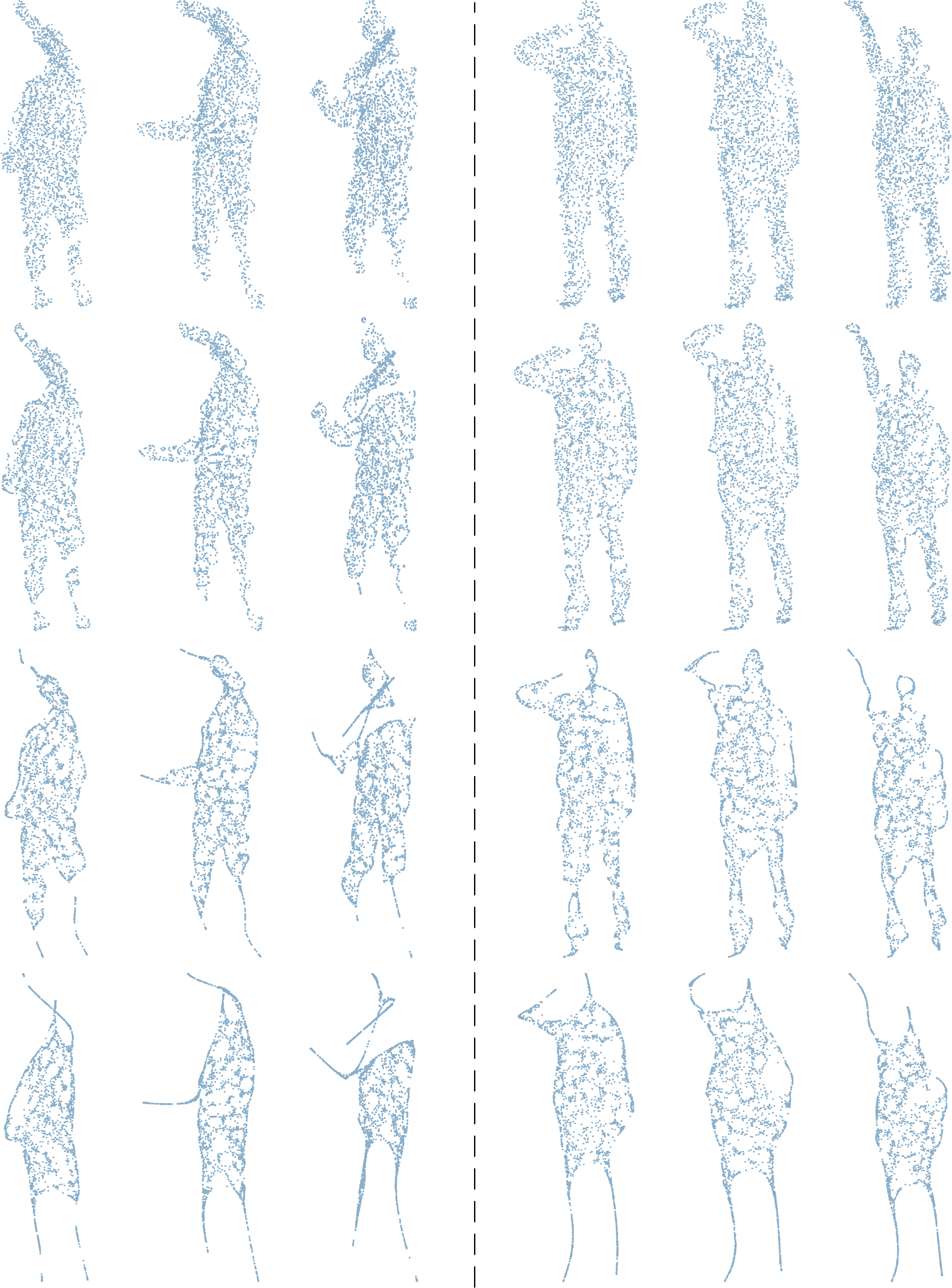}
\caption
{
Visualization of point clouds. 
The first row shows the original point cloud, and the subsequent rows show the reconstructed point clouds where high-frequency bands are progressively removed. 
}
\label{fig:reconstructions}
\end{figure}

Specifically, given the input point cloud video $\left[P; F\right]$, 
we first apply a point 4D convolution \cite{fan2021point} to extract local spatiotemporal features and perform point subsampling, resulting in transformed features $\left[P^{\prime}; F^{\prime}\right]$. 
Next, we construct a $K$-NN graph and transform the sampled point $P^{\prime}$ into the spectral domain via GFT, which can be calculated by:
\begin{equation}
    \hat{{P^{\prime}}} = \operatorname{GFT}({P^{\prime}}). 
\end{equation}
To decompose the signal, we apply a set of spectral filters that partition the spectral domain into three distinct frequency bands: low-frequency band $B_l \in [0,f_l)$, mid-frequency band $B_m \in [f_l,f_h)$, and high-frequency band $B_h \in [f_h,f_{\text{max}}]$. 
This yields the corresponding low-, mid-, and high-frequency GFT coefficients, denoted as $\hat{P}^{\prime}_l$, $\hat{P}^{\prime}_m$, and $\hat{P}^{\prime}_h$, respectively.
The low-frequency components typically capture the coarse global geometric structures of point clouds, while high-frequency components encode fine-grained geometric details such as subtle motion patterns and boundary information, which are very useful in action recognition and semantic segmentation. 

Subsequently, we separately transform the GFT coefficients back to the data domain via IGFT:
\begin{align}
  \begin{split}
  P^{\prime}_{l} &= \operatorname{IGFT}(\hat{{P_{l}^{\prime}}}), \\
  P^{\prime}_{m} &= \operatorname{IGFT}(\hat{{P_{m}^{\prime}}}), \\
 P^{\prime}_{h} &= \operatorname{IGFT}(\hat{{P_{h}^{\prime}}}),
\end{split}
\end{align}
where $P^{\prime}_{l}$, $P^{\prime}_{m}$, and $P^{\prime}_{h}$ are low-, mid-, and high-frequency point clouds, respectively. 

These diverse band-specific reconstructions capture information ranging from global coarse structures to local fine-grained details, thereby enabling a comprehensive understanding of the geometric cues of point clouds. 

We provide visualizations of the original point cloud and its reconstructions after progressively removing high-frequency bands in Figure~\ref{fig:reconstructions}. 
Only results with low-frequency bands are visualized, as high-frequency bands mainly encode high-level semantics that are hard to interpret visually and often yield chaotic, unstructured reconstructions \cite{liu2023point}. 
To better understand the content encoded in high-frequency signals, we can remove them and observe the resulting changes, as shown in Figures~\ref{fig:moti} and~\ref{fig:reconstructions}. 

\subsection{Spatio-Temporal-Spectral Mixer Blocks}

To fully exploit the complementary information across multi-frequency point cloud video, we propose the Spatio-Temporal-Spectral Mixer (STS-Mixer) blocks, as shown in Figure~\ref{fig:overview} (right). 
These blocks are designed not only to refine the feature representations within each frequency band but also to enable information interactions across multiple frequency bands. 
Specifically, the point features $F^{\prime}$, along with multi-frequency point clouds $(X_{l} = \left[P^{\prime}_{l}; F^{\prime}\right], X_{m} =\left[P^{\prime}_{m}; F^{\prime}\right], X_{h} = \left[P^{\prime}_{h}; F^{\prime}\right])$, are fed into the STS-Mixer blocks to model both intra-band spatiotemporal dependencies (\textit{i.e.}, capturing patterns within a given frequency band) and inter-band spectral correlations (\textit{i.e.}, modeling interactions between frequency components). 
Each STS-Mixer block consists of a Frequency-Aware Attention (FA-Attention) module and a Frequency-Mixing MLP (FM-MLP) module. 
These components are described in detail below. 

\noindent \textbf{Frequency-Aware Attention.}
The FA-Attention module is designed to refine features within each frequency band by explicitly modeling their internal spatiotemporal relationships. 
To achieve this, we adopt the Spatio-Temporal Attention (ST-Attention) \cite{fan2022point} as the basic attention block, given its effectiveness in encoding the spatiotemporal structure of point cloud video. 
For each frequency band, we apply a dedicated ST-Attention module, ensuring that the unique spatiotemporal patterns of each band are independently captured without interference from the other bands. 
The update rule for each frequency band at the $\ell$-th STS-Mixer block is expressed as:
\begin{align}
\begin{split}
X_{l}^{\prime \ell} &= {X}_{l}^{\ell-1}+\operatorname{ST-Attention}_{l}(\operatorname{LN}({X}^{\ell-1}_{l})), \\
X_{m}^{\prime \ell} &= {X}_{m}^{\ell-1}+\operatorname{ST-Attention}_{m}(\operatorname{LN}({X}^{\ell-1}_{m})), \\
X_{h}^{\prime \ell} &= {X}_{h}^{\ell-1}+\operatorname{ST-Attention}_{h}(\operatorname{LN}({X}^{\ell-1}_{h})), \\
\end{split}
\end{align}
where $\ell \in\left[1, \ldots, L\right]$ indicates the block index of STS-Mixer blocks, and the initial input to the first block is defined as ${X}^{0}_{l}={X}_{l}$, ${X}^{0}_{m}={X}_{m}$, and ${X}^{0}_{h}={X}_{h}$. 
This frequency-specific attention mechanism allows the model to refine the representation of each frequency band while preserving the different geometric characteristics of each band (\textit{e.g.}, global and local). 

\noindent \textbf{Frequency-Mixing MLP.}
While FA-Attention ensures high-quality intra-frequency modeling, it lacks mechanisms for inter-frequency interaction, which is crucial for understanding the holistic structure of point cloud videos. 
To address this, we introduce the Frequency-Mixing MLP (FM-MLP) to enable information fusion across different frequency point clouds. 
Specifically, the refined features from each frequency branch are concatenated along the channel dimension to form a unified representation. 
This joint representation is then passed through the FM-MLP to allow feature mixing and nonlinear transformation. 
Next, the output ($Y^{l}$) is split back into three frequency bands by evenly partitioning the channels ($X_{l}^{\ell},X_{m}^{\ell},X_{h}^{\ell}$), thus enabling cross-band knowledge propagation. 
The process is described as:
\begin{align}
\begin{split}
\label{equ:mmlp_smt}
X^{\ell} &= \operatorname{Concat}(X_{l}^{\prime \ell},X_{m}^{\prime \ell},X_{h}^{\prime \ell}), \\
Y^{\ell} &= X^{\ell} + \operatorname{FM-MLP}(\operatorname{LN}(X^{\ell})),
\end{split}
\end{align}
where $\operatorname{Concat}(\cdot)$ denotes the concatenation operation, and $\operatorname{FM\text{-}MLP}(\cdot)$ represents the FM-MLP function that follows the standard MLP implementation (Eq.~\ref{equ:mlp}). 
By enabling frequency-wise feature mixing, FM-MLP serves as a bridge that complements the frequency-specific refinement of FA-Attention, ensuring that the final representation captures both intra-band precision and inter-band synergy. 

After several STS-Mixer blocks are stacked, the resulting frequency-aware and spatiotemporal enriched features are concatenated and fed into an MLP to produce point-wise output predictions, such as action classes or semantic labels. 

\begin{table}[t]
\small
\centering
\caption
{
Comparisons of 3D action recognition on the benchmark MSR-Action3D. 
}
\setlength{\tabcolsep}{6.50mm} 
\begin{tabular}{lcc}
\toprule [1pt]
\textbf{Method} &Accuracy \\
\midrule [0.5pt]

MeteorNet (ICCV'19) \cite{liu2019meteornet} &88.50 \\
P4Transformer (CVPR'21) \cite{fan2021point} &90.94 \\
PSTNet (ICLR'21) \cite{fan2022pstnet} &91.20 \\
PSTNet++ (TPAMI'21) \cite{fan2021deep} &92.68 \\
Kinet (CVPR'22) \cite{zhong2022no} &93.27 \\
PST-Transformer (TPAMI'23) \cite{fan2022point} &93.73 \\ 
3DinAction (CVPR'24) \cite{ben20243dinaction} &92.23 \\
Mamba4D (CVPR'25) \cite{liu2024mamba4d}  &93.38 \\
UST-SSM (ICCV'25) \cite{li2025ust}  &94.77 \\

\midrule [0.5pt]

STS-Mixer (Ours)  &\textbf{95.85}\\
  
\bottomrule [1pt]
\end{tabular}
\label{table:msr}
\end{table}

\section{Experiments}

\subsection{Datasets}
We evaluate our method on two widely used datasets: MSR-Action3D \cite{li2010action} and Synthia4D \cite{choy20194d}. 
MSR-Action3D is a 3D action recognition dataset consisting of 567 depth videos captured by Kinect v1, totaling 23k frames.
It has 10 subjects performing 20 distinct action categories. 
Following previous works \cite{shen2023masked,han2024masked}, 270 videos are used for training, and 297 videos are used for testing. 
Synthia4D is a 4D semantic segmentation dataset in outdoor autonomous driving scenarios, built on the Synthia dataset \cite{ros2016synthia} to generate 3D video sequences. 
It includes six driving sequences under nine different weather conditions with moving objects and cameras. 
Following \cite{fan2021point,fan2022point}, we use the same training/validation/test split, consisting of 19,888 frames for training, 815 frames for validation, and 1,886 frames for testing.

\subsection{Implementation Details}
In our implementation, the proposed STS-Mixer consists of $L=3$ stacked STS-Mixer layers with a channel size of $C=128$. 
The model is trained on a single NVIDIA RTX 4090 GPU for 50 epochs using the SGD optimizer. 
The initial learning rate is set to 0.01 and is reduced by a factor of 0.1 at the 20-th and 30-th epochs, respectively. 
We construct a $K$-NN graph with $K = 10$. 
Following \cite{fan2022point}, we set $\{T=24, N=2048, T^{\prime}=12, N^{\prime}=64\}$ for MSR-Action3D and $\{T=3, N=16384, T^{\prime}=3, N^{\prime}=128\}$ for Synthia4D. 
The low- and high-frequency thresholds are set to $f_l=6$ and $f_h=10$. 
Point color information is not used for MSR-Action3D, as the dataset does not provide corresponding color data, but it is utilized for Synthia4D. 

\begin{table}[t]
\small
\centering
\caption
{
Comparisons of 4D semantic segmentation on the benchmark Synthia4D. 
}
\setlength{\tabcolsep}{6.40mm}
\begin{tabular}{lc}
\toprule [1pt]
\textbf{Method} &mIoU (\%) \\

\midrule [0.5pt]  
4D MinkNet14 (CVPR'19) \cite{choy20194d} &77.46 \\
PointNet++ (NeurIPS'17) \cite{qi2017pointnetplus} &79.35 \\
MeteorNet (ICCV'19) \cite{liu2019meteornet} &81.47 \\
PSTNet (ICLR'21) \cite{fan2022pstnet} &82.24 \\
P4Transformer (CVPR'21) \cite{fan2021point} &83.16 \\
PST-Transformer (TPAMI'23) \cite{fan2022point} &{83.95} \\ 
Mamba4D (CVPR'25) \cite{liu2024mamba4d} &83.35 \\
UST-SSM (ICCV'25) \cite{li2025ust} &84.06 \\

\midrule [0.5pt]

STS-Mixer (Ours) &\textbf{84.33} \\

\bottomrule [1pt]
\end{tabular}
\label{table:syn}
\end{table}

\subsection{Comparison with State-of-the-Art Methods}

\paragraph{Results on MSR-Action3D.}
The performance comparisons with state-of-the-art (SOTA) methods on MSR-Action3D are presented in Table~\ref{table:msr}. 
We utilize the best results reported by each work. 
As observed, our proposed method achieves an outstanding action recognition accuracy of 95.85\%, significantly surpassing all previous SOTA approaches.
Notably, STS-Mixer outperforms the recent Mamba4D \cite{liu2024mamba4d} by a margin of 2.47\% in accuracy (95.85\% vs. 93.38\%). 

\begin{table*}[t]
\small
\centering
\caption
{
Ablation study on different components of the proposed STS-Mixer. 
}
\setlength{\tabcolsep}{8.40mm}
\begin{tabular}{l|ccc|c}
\toprule [1pt]
Method& GFT& FA-Attention& FM-MLP& Accuracy (\%) \\
\midrule [0.5pt]
Baseline \cite{fan2022point} & & & & 94.11 \\
Baseline w. GFT& \cmark & & &94.70 \\
STS-Mixer w/o GFT & &\cmark &\cmark &94.50 \\
STS-Mixer w/o FM-MLP & \cmark &\cmark & &95.12 \\
STS-Mixer w/o FA-Attention & \cmark & &\cmark &95.37 \\
STS-Mixer (Ours)& \cmark &\cmark &\cmark &\textbf{95.85}  \\

\bottomrule [1pt]
\end{tabular}
% }
\label{table:component}
\end{table*}

\begin{figure*}[t]
\centering
\includegraphics[width=1.00\linewidth]{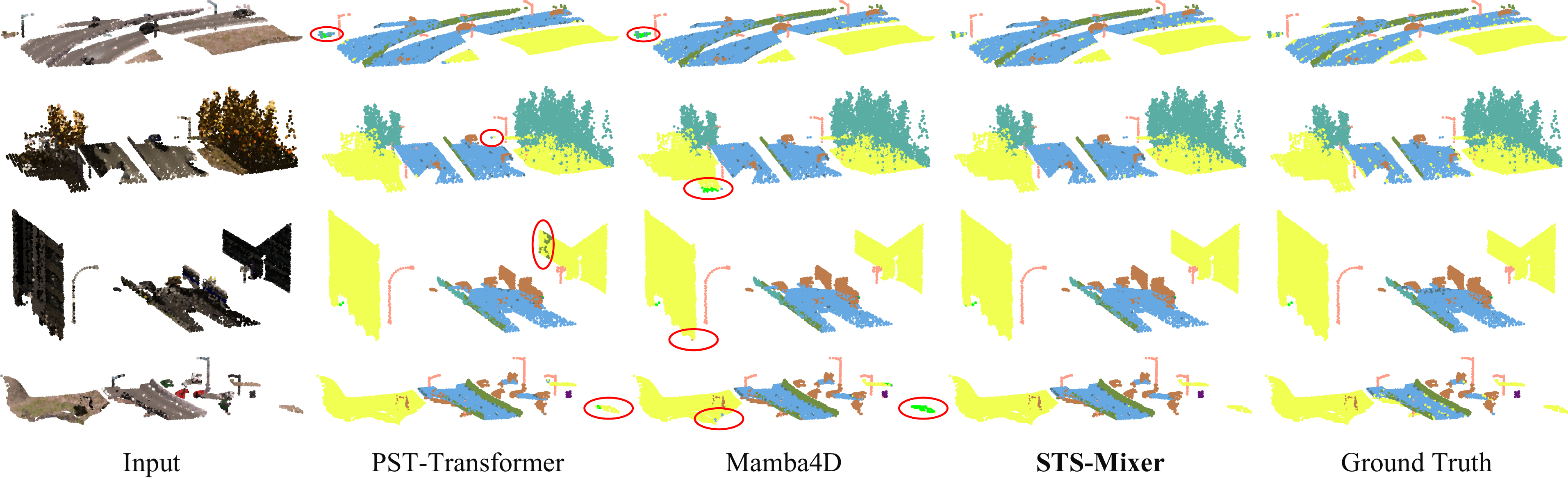}
\caption
{
Qualitative comparison with state-of-the-art methods (PST-Transformer \cite{fan2022point} and Mamba4D \cite{liu2024mamba4d}) on Synthia4D. 
Wrong estimations are highlighted by red circles. 
}
\label{fig:sota}
\end{figure*}

\paragraph{Results on Synthia4D.}
Beyond action recognition, we also evaluate the generalization capability of STS-Mixer on Synthia4D semantic segmentation dataset, with the results shown in Table~\ref{table:syn}. 
STS-Mixer achieves the best mean Intersection-over-Union (mIoU) score of 84.33\%, outperforming strong baselines such as PST-Transformer \cite{fan2022point} (83.95\%) and Mamba4D \cite{liu2024mamba4d} (83.35\%). 
Figure~\ref{fig:sota} presents qualitative comparisons on several complex scenarios. 
It can be seen that STS-Mixer captures more fine-grained details and produces more accurate predictions, validating the benefits of our proposed spatio-temporal-spectral design. 
These consistent improvements across both datasets highlight the robustness and versatility of our approach in handling various 4D point cloud understanding tasks.

\subsection{Ablation Study}

For in-depth analysis, we conduct extensive ablative studies on MSR-Action3D dataset. 

\paragraph{Model Components.}
We also investigate the effectiveness of each component in our design, including the GFT, FA-Attention, and FM-MLP. 
Table~\ref{table:component} presents the ablation results. 
The first row shows the performance of the baseline model \cite{fan2022point}, which does not incorporate spectral representations or any of our proposed components, achieving an accuracy of 94.11\%. 
When introducing either the spectral representation via GFT (Baseline w. GFT) or our architectural modules (FA-Attention and FM-MLP, third row)  into the baseline, the accuracy slightly improves to 94.70\% and 94.50\%, respectively. 
Adding FA-Attention or FM-MLP on top of the baseline with spectral representations leads to further improvements, reaching 95.12\% and 95.37\%, respectively. 
These results demonstrate that each proposed module contributes to performance improvements individually, with the spectral representation playing a particularly significant role. 
The best result is achieved when all components are combined with the baseline (last row), yielding a significant improvement of 1.74\% over the baseline. 
This indicates that the proposed components are complementary and work synergistically to enhance point cloud video understanding. 

\begin{table}[t]
  \small
  \centering
  \caption
  {
Ablation study on different frequency bands. 
  }
  \setlength{\tabcolsep}{2.85mm} 
  \begin{tabular}{lc}
    \toprule [1pt]
    Method &Accuracy (\%) \\
    \midrule [0.5pt]
    STS-Mixer w/o Mid and High Frequency &94.42 \\
    STS-Mixer w/o Low and High Frequency &94.07 \\
    STS-Mixer w/o Low and Mid Frequency &93.72 \\
    STS-Mixer w/o Low Frequency &94.34 \\
    STS-Mixer w/o Mid Frequency &94.77 \\
    STS-Mixer w/o High Frequency &95.12 \\
    STS-Mixer &\textbf{95.85} \\
    \bottomrule [1pt]
  \end{tabular}
  \label{table:frequency_band}
\end{table}

\begin{table}[t]
\small
\centering
\caption
{
Experiments on frequency bands, where $f_l$ and $f_h$ denote the low-frequency threshold and high-frequency threshold, respectively.
}
\setlength{\tabcolsep}{13.90mm} 
\begin{tabular}{cc}
  \toprule [1pt]
  $(f_l, f_h)$ &Accuracy (\%) \\
  \midrule [0.5pt]

  (2, 10) &93.91 \\
  (4, 10) &95.16 \\
  (6, 10) &\textbf{95.85} \\
  (8, 10) &95.33 \\

  \midrule [0.5pt]

  (6, 8) &95.26 \\
  (6, 10) &\textbf{95.85} \\
  (6, 12) &95.29 \\
  (6, 14) &92.96 \\

  \bottomrule [1pt]
\end{tabular}
\label{table:frequency}
\end{table}

\paragraph{Frequency Bands.}
We conduct an ablation study to assess the contribution of each frequency band to the overall performance. 
As shown in Table~\ref{table:frequency_band}, removing any single frequency band leads to a performance drop, indicating that all three bands provide complementary information. 
Notably, excluding the low-frequency band results in the most significant decrease in accuracy (from 95.85\% to 94.34\%), highlighting its critical role in capturing the global structure of point clouds.

\begin{figure}[tb]
\includegraphics[width=1.0\linewidth]
{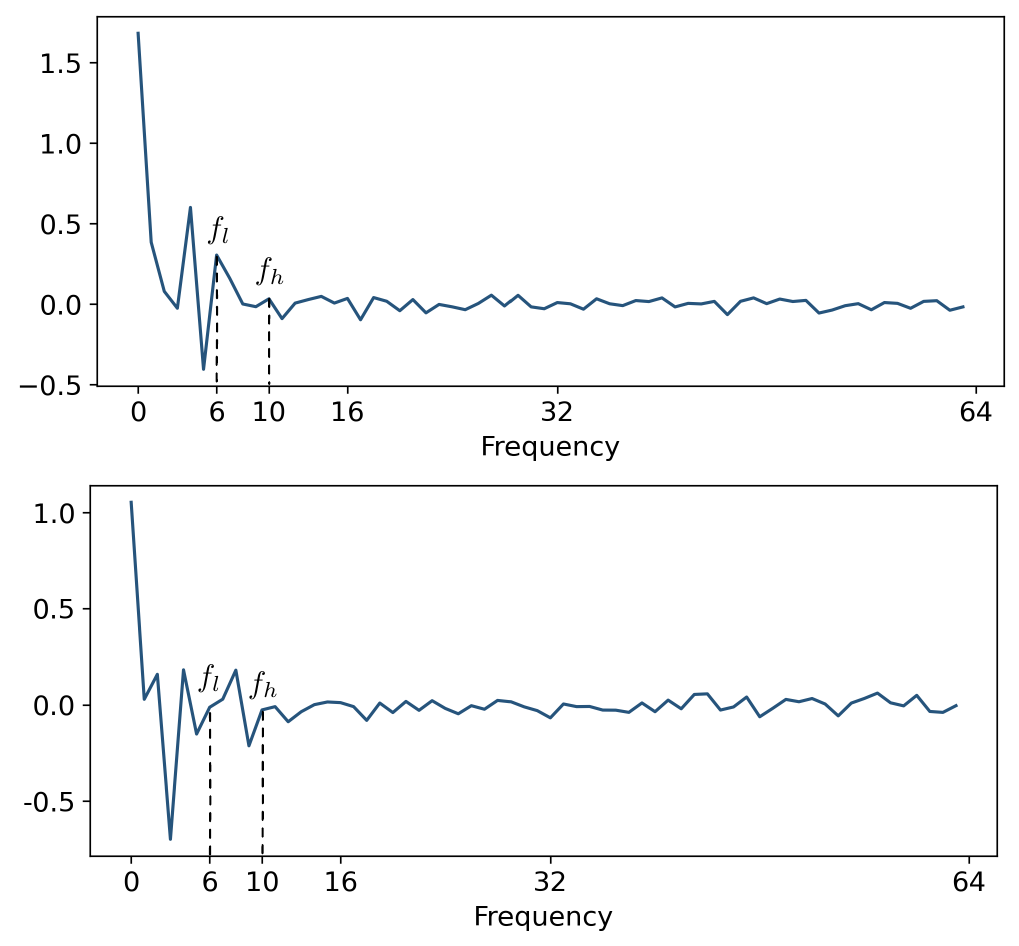}
\caption
{
Visualization of frequency band division.
}
\label{fig:frequency}
\end{figure}

We also conduct an ablation study on different frequency bands by varying the low- and high-frequency thresholds.
The results are presented in Table~\ref{table:frequency}. 
Our STS-Mixer achieves the best performance with the optimal hyperparameters set to $f_l=6$ and $f_h=10$. 
Figure~\ref{fig:frequency} illustrates the division of frequency bands, showing that $f_l = 6$ and $f_h = 10$ lead to a well-balanced energy distribution, where the low- and mid-frequency ranges capture the majority of the energy \cite{hu2022exploring}. 
These thresholds effectively maintain important structural information (low-frequency components) while preserving fine-grained details (high-frequency components), enabling the model to better capture both global context and local variations. 
We also empirically found that adaptive frequency band division yields negligible additional improvements; therefore, we adopt a fixed threshold for the sake of simplicity and efficiency. 

\paragraph{Model Hyperparameters.}
Table~\ref{table:hyperparameter} presents an ablation study on two hyperparameters: the number of STS-Mixer blocks ($L$) and the channel size ($C$).
With $L=3$, increasing $C$ from 64 to 128 significantly boosts accuracy from 93.48\% to 95.85\%. 
However, further increasing $C$ leads to a decrease in performance, despite the increase in model complexity, which suggests that the model may be overfitting. 
Similarly, when fixing $C=128$, increasing $L$ beyond 3 offers limited or no accuracy gains while incurring additional computational costs.
Therefore, the optimal hyperparameter configurations for our model are $L=3$ and $C=128$. 

\begin{table}[t]
\small
\centering
\caption
{
  Experiments on the involved hyperparameters including the number of STS-Mixer blocks $L$ and the number of channels $C$. 
}
\setlength{\tabcolsep}{2.80mm} 
\begin{tabular}{ccccc}
\toprule [1pt]
$L$& $C$ &Params (M) & FLOPs (G) &Accuracy (\%) \\
\midrule [0.5pt]
3& 64& 0.62& 0.95&93.48 \\
3& 128& 2.47& 3.72&\textbf{95.85} \\ 
3& 256& 9.86& 14.69&93.73 \\
3& 512& 39.39& 58.37&92.26 \\
3& 1024& 157.42& 232.70&93.42 \\
\midrule [0.5pt]

2& 128& 1.68& 2.51&94.36 \\
3& 128& 2.47& 3.72&\textbf{95.85} \\
4& 128& 3.26& 4.93&94.50 \\
5& 128& 4.05& 6.15&94.81 \\
6& 128& 4.84& 7.36&94.84\\

\bottomrule [1pt]
\end{tabular}
\label{table:hyperparameter}
\end{table}

\paragraph{Number $K$.}
In Table~\ref{table:number_k}, we investigate the effect of varying the neighbor number $K$ in the $K$-NN graph. 
The $K$-NN graph plays a critical role in capturing local geometric relationships among points, which directly influences the effectiveness of spectral representation learning. 
The results show that as $K$ increases from 6 to 10, the accuracy improves, reaching a peak of 95.85\% at $K=10$. 
This is because expanding the local neighborhood enriches the geometric context, thereby enhancing feature representation. 
However, further increasing $K$ beyond 10 results in a performance drop. 
This is likely because a moderate number of neighbors provides a good balance between capturing local geometric structures and avoiding noise or irrelevant connections introduced by overly large neighborhoods.

\begin{table}[t]
  \small
  \centering
  \caption
  { 
      Experiments on the neighbor number $K$. 
  }  
  \setlength{\tabcolsep}{1.65mm}
  \begin{tabular}{cccccccc}
  \toprule [1pt]
  $K$ &6 &8 &10 &12 &14 &16 \\
  \midrule [0.5pt]
  Accuracy (\%)& 93.94 &94.74 &\textbf{95.85} &93.84 &93.73 &93.38 \\
  \bottomrule [1pt]
  \end{tabular}
  \label{table:number_k}
\end{table}

\section{Conclusion}
This paper presents Spatio-Temporal-Spectral Mixer (STS-Mixer), a unified framework that combines spatiotemporal and spectral representations for effective 4D point cloud video understanding. 
By transforming point clouds into the graph spectral domain, STS-Mixer effectively decomposes them into multiple frequency components to capture distinct geometric information from global structures and fine-grained details. 
The proposed frequency-aware modules, FA-Attention and FM-MLP, further refine and integrate multi-frequency representations with spatiotemporal cues, while enabling fine-grained and holistic representation learning of point cloud videos. 
Extensive experiments on 3D action recognition and 4D semantic segmentation benchmarks demonstrate that STS-Mixer consistently outperforms state-of-the-art methods. 

{\small
\noindent \textbf{Acknowledgements.}
This study is funded by the Ministry of Education Singapore, under the Tier-2 project scheme with project number MOET2EP20123-0003.
}

{
  \small
  \bibliographystyle{ieeenat_fullname}
  \bibliography{refs}

\begin{thebibliography}{48}
\providecommand{\natexlab}[1]{#1}
\providecommand{\url}[1]{\texttt{#1}}
\expandafter\ifx\csname urlstyle\endcsname\relax
  \providecommand{\doi}[1]{doi: #1}\else
  \providecommand{\doi}{doi: \begingroup \urlstyle{rm}\Url}\fi

\bibitem[Ba et~al.(2016)Ba, Kiros, and Hinton]{ba2016layer}
Jimmy~Lei Ba, Jamie~Ryan Kiros, and Geoffrey~E Hinton.
\newblock Layer normalization.
\newblock \emph{arXiv preprint arXiv:1607.06450}, 2016.

\bibitem[Ben-Shabat et~al.(2024)Ben-Shabat, Shrout, and Gould]{ben20243dinaction}
Yizhak Ben-Shabat, Oren Shrout, and Stephen Gould.
\newblock 3{D}in{A}ction: Understanding human actions in 3{D} point clouds.
\newblock In \emph{CVPR}, pages 19978--19987, 2024.

\bibitem[Chen et~al.(2017)Chen, Tian, Feng, Vetro, and Kova{\v{c}}evi{\'c}]{chen2017fast}
Siheng Chen, Dong Tian, Chen Feng, Anthony Vetro, and Jelena Kova{\v{c}}evi{\'c}.
\newblock Fast resampling of three-dimensional point clouds via graphs.
\newblock \emph{IEEE Transactions on Signal Processing}, 66\penalty0 (3):\penalty0 666--681, 2017.

\bibitem[Choe et~al.(2022)Choe, Park, Rameau, Park, and Kweon]{choe2022pointmixer}
Jaesung Choe, Chunghyun Park, Francois Rameau, Jaesik Park, and In~So Kweon.
\newblock Point{M}ixer: {MLP}-{M}ixer for point cloud understanding.
\newblock In \emph{ECCV}, pages 620--640, 2022.

\bibitem[Choy et~al.(2019)Choy, Gwak, and Savarese]{choy20194d}
Christopher Choy, JunYoung Gwak, and Silvio Savarese.
\newblock 4{D} spatio-temporal convnets: Minkowski convolutional neural networks.
\newblock In \emph{CVPR}, pages 3075--3084, 2019.

\bibitem[Dosovitskiy et~al.(2021)Dosovitskiy, Beyer, Kolesnikov, Weissenborn, Zhai, Unterthiner, Dehghani, Minderer, Heigold, Gelly, Uszkoreit, and Houlsby]{vit}
Alexey Dosovitskiy, Lucas Beyer, Alexander Kolesnikov, Dirk Weissenborn, Xiaohua Zhai, Thomas Unterthiner, Mostafa Dehghani, Matthias Minderer, Georg Heigold, Sylvain Gelly, Jakob Uszkoreit, and Neil Houlsby.
\newblock An image is worth 16x16 words: Transformers for image recognition at scale.
\newblock In \emph{ICLR}, 2021.

\bibitem[Duan et~al.(2019)Duan, Zheng, Lu, Zhou, and Tian]{duan2019structural}
Yueqi Duan, Yu Zheng, Jiwen Lu, Jie Zhou, and Qi Tian.
\newblock Structural relational reasoning of point clouds.
\newblock In \emph{CVPR}, pages 949--958, 2019.

\bibitem[Fan and Yang(2019)]{fan2019pointrnn}
Hehe Fan and Yi Yang.
\newblock Point{RNN}: Point recurrent neural network for moving point cloud processing.
\newblock \emph{arXiv preprint arXiv:1910.08287}, 2019.

\bibitem[Fan et~al.(2021{\natexlab{a}})Fan, Yang, and Kankanhalli]{fan2021point}
Hehe Fan, Yi Yang, and Mohan Kankanhalli.
\newblock Point 4{D} transformer networks for spatio-temporal modeling in point cloud videos.
\newblock In \emph{CVPR}, pages 14204--14213, 2021{\natexlab{a}}.

\bibitem[Fan et~al.(2021{\natexlab{b}})Fan, Yu, Ding, Yang, and Kankanhalli]{fan2022pstnet}
Hehe Fan, Xin Yu, Yuhang Ding, Yi Yang, and Mohan Kankanhalli.
\newblock {PSTN}et: Point spatio-temporal convolution on point cloud sequences.
\newblock In \emph{ICLR}, 2021{\natexlab{b}}.

\bibitem[Fan et~al.(2021{\natexlab{c}})Fan, Yu, Yang, and Kankanhalli]{fan2021deep}
Hehe Fan, Xin Yu, Yi Yang, and Mohan Kankanhalli.
\newblock Deep hierarchical representation of point cloud videos via spatio-temporal decomposition.
\newblock \emph{IEEE Transactions on Pattern Analysis and Machine Intelligence}, 44\penalty0 (12):\penalty0 9918--9930, 2021{\natexlab{c}}.

\bibitem[Fan et~al.(2023)Fan, Yang, and Kankanhalli]{fan2022point}
Hehe Fan, Yi Yang, and Mohan Kankanhalli.
\newblock Point spatio-temporal transformer networks for point cloud video modeling.
\newblock \emph{IEEE Transactions on Pattern Analysis and Machine Intelligence}, 45\penalty0 (2):\penalty0 2181--2192, 2023.

\bibitem[Hammond et~al.(2011)Hammond, Vandergheynst, and Gribonval]{hammond2011wavelets}
David~K Hammond, Pierre Vandergheynst, and R{\'e}mi Gribonval.
\newblock Wavelets on graphs via spectral graph theory.
\newblock \emph{Applied and Computational Harmonic Analysis}, 30\penalty0 (2):\penalty0 129--150, 2011.

\bibitem[Han et~al.(2024)Han, Xu, Xu, Qian, and Xie]{han2024masked}
Yuehui Han, Can Xu, Rui Xu, Jianjun Qian, and Jin Xie.
\newblock Masked motion prediction with semantic contrast for point cloud sequence learning.
\newblock In \emph{ECCV}, pages 414--431, 2024.

\bibitem[Hu et~al.(2022)Hu, Liu, and Hu]{hu2022exploring}
Qianjiang Hu, Daizong Liu, and Wei Hu.
\newblock Exploring the devil in graph spectral domain for 3{D} point cloud attacks.
\newblock In \emph{ECCV}, pages 229--248, 2022.

\bibitem[Hu et~al.(2021)Hu, Pang, Liu, Tian, Lin, and Vetro]{hu2021graph}
Wei Hu, Jiahao Pang, Xianming Liu, Dong Tian, Chia-Wen Lin, and Anthony Vetro.
\newblock Graph signal processing for geometric data and beyond: Theory and applications.
\newblock \emph{IEEE Transactions on Multimedia}, 24:\penalty0 3961--3977, 2021.

\bibitem[Jing et~al.(2024)Jing, Xue, Yan, Zheng, Wang, Zhang, Wang, Fang, Zhao, and Li]{jing2024x4d}
Linglin Jing, Ying Xue, Xu Yan, Chaoda Zheng, Dong Wang, Ruimao Zhang, Zhigang Wang, Hui Fang, Bin Zhao, and Zhen Li.
\newblock X4{D}-{S}ceneformer: Enhanced scene understanding on 4{D} point cloud videos through cross-modal knowledge transfer.
\newblock In \emph{AAAI}, pages 2670--2678, 2024.

\bibitem[Li et~al.(2025)Li, Wang, Yuan, Liu, Meng, Yuan, and Liu]{li2025ust}
Peiming Li, Ziyi Wang, Yulin Yuan, Hong Liu, Xiangming Meng, Junsong Yuan, and Mengyuan Liu.
\newblock {UST}-{SSM}: Unified spatio-temporal state space models for point cloud video modeling.
\newblock In \emph{ICCV}, 2025.

\bibitem[Li et~al.(2010)Li, Zhang, and Liu]{li2010action}
Wanqing Li, Zhengyou Zhang, and Zicheng Liu.
\newblock Action recognition based on a bag of 3{D} points.
\newblock In \emph{CVPR Workshops}, pages 9--14, 2010.

\bibitem[Li et~al.(2022)Li, Liu, Tang, Wang, and Van~Gool]{mhformer}
Wenhao Li, Hong Liu, Hao Tang, Pichao Wang, and Luc Van~Gool.
\newblock {MHF}ormer: Multi-hypothesis transformer for 3{D} human pose estimation.
\newblock In \emph{CVPR}, pages 13147--13156, 2022.

\bibitem[Li et~al.(2024)Li, Liu, Liu, Wang, Cai, and Sebe]{hot}
Wenhao Li, Mengyuan Liu, Hong Liu, Pichao Wang, Jialun Cai, and Nicu Sebe.
\newblock Hourglass tokenizer for efficient transformer-based 3{D} human pose estimation.
\newblock In \emph{CVPR}, pages 604--613, 2024.

\bibitem[Liang et~al.(2024{\natexlab{a}})Liang, Feng, Zhou, Zhang, Zou, and Bai]{liang2024parameter}
Dingkang Liang, Tianrui Feng, Xin Zhou, Yumeng Zhang, Zhikang Zou, and Xiang Bai.
\newblock Parameter-efficient fine-tuning in spectral domain for point cloud learning.
\newblock \emph{arXiv preprint arXiv:2410.08114}, 2024{\natexlab{a}}.

\bibitem[Liang et~al.(2024{\natexlab{b}})Liang, Zhou, Xu, Zhu, Zou, Ye, Tan, and Bai]{liang2024pointmamba}
Dingkang Liang, Xin Zhou, Wei Xu, Xingkui Zhu, Zhikang Zou, Xiaoqing Ye, Xiao Tan, and Xiang Bai.
\newblock Point{M}amba: A simple state space model for point cloud analysis.
\newblock In \emph{NeurIPS}, 2024{\natexlab{b}}.

\bibitem[Liu et~al.(2023{\natexlab{a}})Liu, Hu, and Li]{liu2023point}
Daizong Liu, Wei Hu, and Xin Li.
\newblock Point cloud attacks in graph spectral domain: When 3{D} geometry meets graph signal processing.
\newblock \emph{IEEE TPAMI}, 46\penalty0 (5):\penalty0 3079--3095, 2023{\natexlab{a}}.

\bibitem[Liu et~al.(2025)Liu, Han, Liu, Aviles-Rivero, Jiang, Liu, and Wang]{liu2024mamba4d}
Jiuming Liu, Jinru Han, Lihao Liu, Angelica~I Aviles-Rivero, Chaokang Jiang, Zhe Liu, and Hesheng Wang.
\newblock Mamba4{D}: Efficient long-sequence point cloud video understanding with disentangled spatial-temporal state space models.
\newblock In \emph{CVPR}, 2025.

\bibitem[Liu et~al.(2019)Liu, Yan, and Bohg]{liu2019meteornet}
Xingyu Liu, Mengyuan Yan, and Jeannette Bohg.
\newblock Meteor{N}et: Deep learning on dynamic 3{D} point cloud sequences.
\newblock In \emph{ICCV}, pages 9246--9255, 2019.

\bibitem[Liu et~al.(2023{\natexlab{b}})Liu, Chen, Zhang, Huang, and Yi]{liu2023leaf}
Yunze Liu, Junyu Chen, Zekai Zhang, Jingwei Huang, and Li Yi.
\newblock Lea{F}: learning frames for 4{D} point cloud sequence understanding.
\newblock In \emph{ICCV}, pages 604--613, 2023{\natexlab{b}}.

\bibitem[Liu et~al.(2021)Liu, Lin, Cao, Hu, Wei, Zhang, Lin, and Guo]{swin}
Ze Liu, Yutong Lin, Yue Cao, Han Hu, Yixuan Wei, Zheng Zhang, Stephen Lin, and Baining Guo.
\newblock Swin {T}ransformer: Hierarchical vision transformer using shifted windows.
\newblock In \emph{ICCV}, pages 10012--10022, 2021.

\bibitem[Ma et~al.(2022)Ma, Qin, You, Ran, and Fu]{ma2022rethinking}
Xu Ma, Can Qin, Haoxuan You, Haoxi Ran, and Yun Fu.
\newblock Rethinking network design and local geometry in point cloud: A simple residual {MLP} framework.
\newblock In \emph{ICLR}, 2022.

\bibitem[Qi et~al.(2017{\natexlab{a}})Qi, Su, Mo, and Guibas]{qi2017pointnet}
Charles~R Qi, Hao Su, Kaichun Mo, and Leonidas~J Guibas.
\newblock Point{N}et: Deep learning on point sets for 3{D} classification and segmentation.
\newblock In \emph{CVPR}, pages 652--660, 2017{\natexlab{a}}.

\bibitem[Qi et~al.(2017{\natexlab{b}})Qi, Yi, Su, and Guibas]{qi2017pointnetplus}
Charles~Ruizhongtai Qi, Li Yi, Hao Su, and Leonidas~J Guibas.
\newblock Point{N}et++: Deep hierarchical feature learning on point sets in a metric space.
\newblock In \emph{NeurIPS}, 2017{\natexlab{b}}.

\bibitem[Ramasinghe et~al.(2020)Ramasinghe, Khan, Barnes, and Gould]{ramasinghe2020spectral}
Sameera Ramasinghe, Salman Khan, Nick Barnes, and Stephen Gould.
\newblock Spectral-{GAN}s for high-resolution 3{D} point-cloud generation.
\newblock In \emph{IROS}, pages 8169--8176, 2020.

\bibitem[Ros et~al.(2016)Ros, Sellart, Materzynska, Vazquez, and Lopez]{ros2016synthia}
German Ros, Laura Sellart, Joanna Materzynska, David Vazquez, and Antonio~M Lopez.
\newblock The {S}ynthia dataset: A large collection of synthetic images for semantic segmentation of urban scenes.
\newblock In \emph{CVPR}, pages 3234--3243, 2016.

\bibitem[Rosman et~al.(2013)Rosman, Dubrovina, and Kimmel]{rosman2013patch}
Guy Rosman, Anastasia Dubrovina, and Ron Kimmel.
\newblock Patch-collaborative spectral point-cloud denoising.
\newblock In \emph{Computer Graphics Forum}, pages 1--12, 2013.

\bibitem[Shen et~al.(2023{\natexlab{a}})Shen, Sheng, Fan, Wang, Guo, Liu, Wen, and Zhou]{shen2023masked}
Zhiqiang Shen, Xiaoxiao Sheng, Hehe Fan, Longguang Wang, Yulan Guo, Qiong Liu, Hao Wen, and Xi Zhou.
\newblock Masked spatio-temporal structure prediction for self-supervised learning on point cloud videos.
\newblock In \emph{ICCV}, pages 16580--16589, 2023{\natexlab{a}}.

\bibitem[Shen et~al.(2023{\natexlab{b}})Shen, Sheng, Wang, Guo, Liu, and Zhou]{shen2023pointcmp}
Zhiqiang Shen, Xiaoxiao Sheng, Longguang Wang, Yulan Guo, Qiong Liu, and Xi Zhou.
\newblock Point{CMP}: Contrastive mask prediction for self-supervised learning on point cloud videos.
\newblock In \emph{CVPR}, pages 1212--1222, 2023{\natexlab{b}}.

\bibitem[Sheng et~al.(2023{\natexlab{a}})Sheng, Shen, and Xiao]{sheng2023contrastive}
Xiaoxiao Sheng, Zhiqiang Shen, and Gang Xiao.
\newblock Contrastive predictive autoencoders for dynamic point cloud self-supervised learning.
\newblock In \emph{AAAI}, pages 9802--9810, 2023{\natexlab{a}}.

\bibitem[Sheng et~al.(2023{\natexlab{b}})Sheng, Shen, Xiao, Wang, Guo, and Fan]{sheng2023point}
Xiaoxiao Sheng, Zhiqiang Shen, Gang Xiao, Longguang Wang, Yulan Guo, and Hehe Fan.
\newblock Point contrastive prediction with semantic clustering for self-supervised learning on point cloud videos.
\newblock In \emph{ICCV}, pages 16515--16524, 2023{\natexlab{b}}.

\bibitem[Shuman et~al.(2013)Shuman, Narang, Frossard, Ortega, and Vandergheynst]{shuman2013emerging}
David~I Shuman, Sunil~K Narang, Pascal Frossard, Antonio Ortega, and Pierre Vandergheynst.
\newblock The emerging field of signal processing on graphs: Extending high-dimensional data analysis to networks and other irregular domains.
\newblock \emph{IEEE Signal Processing Magazine}, 30\penalty0 (3):\penalty0 83--98, 2013.

\bibitem[Wang et~al.(2025)Wang, Xu, Ding, Zhang, Bai, and Li]{wangpvnext}
Jie Wang, Tingfa Xu, Lihe Ding, Xinjie Zhang, Long Bai, and Jianan Li.
\newblock Pv{N}e{X}t: Rethinking network design and temporal motion for point cloud video recognition.
\newblock In \emph{ICLR}, 2025.

\bibitem[Wen et~al.(2024)Wen, Long, Yu, and Tao]{wen2024pointwavelet}
Cheng Wen, Jianzhi Long, Baosheng Yu, and Dacheng Tao.
\newblock Point{W}avelet: Learning in spectral domain for 3-d point cloud analysis.
\newblock \emph{IEEE Transactions on Neural Networks and Learning Systems}, 2024.

\bibitem[Wen et~al.(2022)Wen, Liu, Huang, Duan, and Yi]{wen2022point}
Hao Wen, Yunze Liu, Jingwei Huang, Bo Duan, and Li Yi.
\newblock Point primitive transformer for long-term 4{D} point cloud video understanding.
\newblock In \emph{ECCV}, pages 19--35, 2022.

\bibitem[Yang et~al.(2019)Yang, Zhang, Ni, Li, Liu, Zhou, and Tian]{yang2019modeling}
Jiancheng Yang, Qiang Zhang, Bingbing Ni, Linguo Li, Jinxian Liu, Mengdie Zhou, and Qi Tian.
\newblock Modeling point clouds with self-attention and gumbel subset sampling.
\newblock In \emph{CVPR}, pages 3323--3332, 2019.

\bibitem[Zhang et~al.(2014)Zhang, Florencio, and Loop]{zhang2014point}
Cha Zhang, Dinei Florencio, and Charles Loop.
\newblock Point cloud attribute compression with graph transform.
\newblock In \emph{ICIP}, pages 2066--2070, 2014.

\bibitem[Zhang et~al.(2020)Zhang, Cui, and Ding]{zhang2020hypergraph}
Songyang Zhang, Shuguang Cui, and Zhi Ding.
\newblock Hypergraph spectral analysis and processing in 3{D} point cloud.
\newblock \emph{IEEE Transactions on Image Processing}, 30:\penalty0 1193--1206, 2020.

\bibitem[Zhang et~al.(2025)Zhang, Yuan, Qi, Zhang, Zhou, Ji, Yan, and Li]{zhang2025point}
Tao Zhang, Haobo Yuan, Lu Qi, Jiangning Zhang, Qianyu Zhou, Shunping Ji, Shuicheng Yan, and Xiangtai Li.
\newblock Point {C}loud {M}amba: Point cloud learning via state space model.
\newblock In \emph{AAAI}, pages 10121--10130, 2025.

\bibitem[Zhang et~al.(2023)Zhang, Dong, Liu, and Yi]{zhang2023complete}
Zhuoyang Zhang, Yuhao Dong, Yunze Liu, and Li Yi.
\newblock Complete-to-partial 4{D} distillation for self-supervised point cloud sequence representation learning.
\newblock In \emph{CVPR}, pages 17661--17670, 2023.

\bibitem[Zhong et~al.(2022)Zhong, Zhou, Hu, Wang, Trigoni, and Markham]{zhong2022no}
Jia-Xing Zhong, Kaichen Zhou, Qingyong Hu, Bing Wang, Niki Trigoni, and Andrew Markham.
\newblock No pain, big gain: classify dynamic point cloud sequences with static models by fitting feature-level space-time surfaces.
\newblock In \emph{CVPR}, pages 8510--8520, 2022.

\end{thebibliography}
}

\end{document}